\documentclass[10pt,twocolumn,letterpaper]{article}

 \usepackage{cvpr}              
\definecolor{cvprblue}{rgb}{0.21,0.49,0.74}
\usepackage[pagebackref,breaklinks,colorlinks,allcolors=cvprblue]{hyperref}
\usepackage{booktabs}
\usepackage{multirow}
\usepackage{pifont}
\usepackage{graphicx}
\usepackage{tabularx}
\usepackage{amsmath}
\usepackage{tcolorbox}
\usepackage{xcolor}
\usepackage{colortbl}
\usepackage{threeparttable}
\usepackage{makecell}
\usepackage{array} 
\usepackage{adjustbox}
\usepackage[accsupp]{axessibility}  

\def\paperID{} 
\def\confName{CVPR}
\def\confYear{2026}

\title{MMGait: Towards Multi-Modal Gait Recognition}

\author{
	Chenye Wan$^{1\dagger}$ \hskip1.6em Qingyuan Cai$^{1\dagger}$ \hskip1.6em Saihui Hou$^{1}$ \hskip1.6em Aoqi Li$^{1}$ \hskip1.6em Yongzhen Huang$^{1，2}$ \vspace{+2mm} \\ 
	$^{1}$School of Artificial Intelligence, Beijing Normal University, $^{2}$WATRIX.AI  \\
	{\tt\small \{chenye.wang, caiqingyuan, rookie\}@mail.bnu.edu.cn \{housaihui,huangyonzhen\}@bnu.edu.cn \\ 
	\small{\url{https://github.com/BNU-IVC/MMGait} \\
	}
}
}
\author{
	Chenye Wang$^{1\dagger}$ \hskip1.6em
	Qingyuan Cai$^{1\dagger}$ \hskip1.6em
	Saihui Hou$^{1*}$ \hskip1.6em
	Aoqi Li$^{1}$ \hskip1.6em
	Yongzhen Huang$^{1,2}$ \vspace{+2mm} \\ 
	$^{1}$School of Artificial Intelligence, Beijing Normal University \hskip1.6em
	$^{2}$WATRIX.AI \\
{\ttfamily\small
	\{chenye.wang, caiqingyuan, rookie\}@mail.bnu.edu.cn 
	\{housaihui, huangyongzhen\}@bnu.edu.cn
}
}

\begin{document}
	\maketitle

 \begin{abstract}
	Gait recognition has emerged as a powerful biometric technique for identifying individuals at a distance without requiring user cooperation.
	Most existing methods focus primarily on RGB-derived modalities, which fall short in real-world scenarios requiring multi-modal collaboration and cross-modal retrieval.
	To overcome these challenges, we present \textbf{MMGait}, a comprehensive multi-modal gait benchmark integrating data from five heterogeneous sensors, including an RGB camera, a depth camera, an infrared camera, a LiDAR scanner, and a 4D Radar system.
	MMGait contains twelve modalities and 334,060 sequences from 725 subjects, enabling systematic exploration across geometric, photometric, and motion domains.
	Based on MMGait, we conduct extensive evaluations on single-modal, cross-modal, and multi-modal paradigms to analyze modality robustness and complementarity.
	Furthermore, we introduce a new task, \textbf{Omni Multi-Modal Gait Recognition}, which aims to unify the above three gait recognition paradigms within a single model. 
	We also propose a simple yet powerful baseline, \textbf{OmniGait}, which learns a shared embedding space across diverse modalities and achieves promising recognition performance.
	The MMGait benchmark, codebase, and pretrained checkpoints are publicly available at \href{https://github.com/BNU-IVC/MMGait}{https://github.com/BNU-IVC/MMGait}.
\end{abstract}

 \begingroup
 \renewcommand\thefootnote{}
 \footnotetext{$\dagger$ Equal contribution. \quad * Corresponding author.}
 \endgroup
 
 \vspace{-4mm}
\section{Introduction}

\begin{figure*}[t]
	\centering
	\includegraphics[width=\linewidth]{figures/overview.pdf}
	\vspace{-6mm}
	\caption{Highlights of MMGait. MMGait integrates data from five sensors. It spans twelve modalities, including multi-view, multi-modal representations, and multi-covariate conditions. }
	\label{Fig:overview}
	\vspace{-5mm}

\end{figure*}
Gait recognition, as a non-contact biometric technique, has attracted growing attention due to its unique capability of identifying individuals from a distance without requiring user cooperation~\cite{sepas2022deep,shen2024comprehensive}. 
In recent years, the community has witnessed remarkable progress, especially with the dominance of RGB-derived modalities such as silhouettes~\cite{xiong2024causality, huang2025learning,yang2025bridging} and pose sequences~\cite{cosma2025model,meng2025fastposegait,guo2025gaitcontour}, which have achieved promising performance in both indoor and outdoor scenarios~\cite{ye2024biggait,jin2025denoising,ye2025biggergait,lang2025beyond}.

However, despite their success, RGB-derived modalities suffer from several inherent limitations. 
They lack 3D perception and are vulnerable to occlusion and challenging conditions (\textit{e.g.}, rain, fog, and low illumination), motivating the exploration of alternative sensing modalities beyond RGB for more robust and generalizable gait representations~\cite{shen2023lidargait,huang2024occluded}.
Fortunately, the emergence of multi-modal gait benchmarks (\textit{e.g.}, LidarGait~\cite{shen2023lidargait} and FreeGait~\cite{han2024gait}) has inspired a growing interest in leveraging LiDAR point clouds. These datasets have encouraged the study of point cloud-based recognition~\cite{wang2023pointgait,guo2024lidar}, cross-modal retrieval~\cite{wang2024cross,guo2024camera}, and multi-modal fusion~\cite{deng2024licaf}. Nevertheless, existing benchmarks are still limited in scope, as they typically include only RGB and LiDAR sensors.
In particular, current datasets limit the study of heterogeneous modality interactions and unified cross-sensor retrieval, both of which are crucial for the deployment of gait recognition systems in practical environments~\cite{liu2024cross,yang2024detection,han2025mmreid}.

To bridge this gap, we introduce \textbf{MMGait}, a comprehensive multi-modal gait dataset designed to advance gait recognition toward realistic multi-sensor settings. MMGait collects data from \textit{five distinct sensors}: an RGB Camera, a LiDAR Scanner, a Depth Camera, an Infrared (IR) camera, and a 4D Radar System. These sensors collectively span the spectrum from \textit{visible light} to \textit{infrared}, from \textit{2D appearance characteristics }to \textit{3D geometric structural information}, and from \textit{low-cost} to \textit{high-cost} devices, providing complementary cues across geometric, photometric, and motion domains.

MMGait possesses several key characteristics, as shown in Figure~\ref{Fig:overview}: (1) \textit{\textbf{Comprehensive}}. It includes data from five heterogeneous sensors, which are processed into twelve modalities, covering widely used representations (\textit{e.g.}, silhouettes, pose) as well as emerging ones (\textit{e.g.}, LiDAR point cloud, event). (2) \textit{\textbf{Precision}}. The dataset is captured using advanced devices and includes multiple walking conditions and view annotations for each identity, enabling rigorous cross-view and cross-covariate evaluation. (3) \textit{\textbf{Scalability}}. MMGait contains 725 subjects, each with over 460 sequences on average, making it one of the most diverse and large-scale multi-modal gait benchmark to date.

Building upon MMGait, we conduct comprehensive benchmark studies across three fundamental perspectives of gait recognition:
(1) \textit{\textbf{Single-Modal Recognition}}, to analyze the potential of each modality under a unified model architecture;
(2) \textit{\textbf{Cross-Modal Recognition}}, to evaluate the generalization and transferability of gait representations across different modalities;
and (3)\textit{ \textbf{Multi-Modal Recognition}}, to explore the complementarity between modalities, both \textit{intra-sensor} (\textit{e.g.}, silhouette + event) and \textit{inter-sensor} (\textit{e.g.}, RGB Camera + LiDAR Scanner).

While these analyses offer deep insights into modality behavior, they also raise a more challenging and practical question:
\textbf{\textit{Can we design a single model that unifies all three paradigms within one framework?}}

To tackle this challenge, we propose a new task termed \textbf{Omni Multi-Modal Gait Recognition}, which aims to develop \textit{a unified model} capable of accepting any types of modality as a query and retrieving targets from any modality. 
This unified design offers remarkable flexibility to accommodate diverse modalities, effectively exploits their complementary information for richer and more discriminative gait representations, and enhances computational efficiency by learning shared feature spaces instead of training separate models.
Such a design greatly improves practicality, enabling seamless deployment in real-world multi-sensor systems under diverse conditions~\cite{zuo2025reid5o, li2024all}.

Building on this task, we further present a simple yet strong baseline, \textbf{OmniGait}, which learns a shared embedding space across diverse modalities. Acting as a versatile and unified model, OmniGait achieves performance comparable to, and in certain cases surpassing, separate models. Its flexibility and balanced performance across heterogeneous modalities suggest that Omni Multi-Modal Gait Recognition represents a promising and practical research direction for the future of multi-modal gait systems.

In summary, our contributions are threefold:
\begin{itemize}
	\item We construct MMGait, one of the most comprehensive multi-modal gait recognition dataset to date, comprising 725 subjects and 334,060 sequences across 12 modalities derived from 5 heterogeneous sensors. 
	\item We conduct detailed evaluations across single-modal, cross-modal, and multi-modal settings to provide a holistic understanding of modality characteristics, robustness, and complementarity.
	\item We propose a challenging task of Omni Multi-Modal Gait Recognition and present a simple yet effective baseline, OmniGait, demonstrating its potential to unify diverse gait recognition paradigms under a single framework.
\end{itemize}

 \section{MMGait Benchmark}

\subsection{Sensor Setup}

To comprehensively cover real-world multi-sensor scenarios, MMGait employs five heterogeneous sensing devices: an RGB Camera, an IR Camera, a Depth Camera, a LiDAR Scanner, and a 4D Radar System. 

\begin{table}[t] 
	\centering \caption{Specifications of sensing devices and corresponding modalities in MMGait.} \label{Tab:sensor} \renewcommand{\arraystretch}{1.4} \resizebox{0.49\textwidth}{!}
	{ \begin{tabular}{ccccc} \toprule Sensor & Range (m) & Resolution  & Modality \\ \midrule RGB Camera & 0.4--10 & 1280×800 & \makecell[c]{RGB, Silhouette, 2D Pose,\\ 3D Pose, Event} \\ \hline IR Camera & 0.4--10 & 1280×700& \makecell[c]{IR, Silhouette} \\ \hline Depth Camera & 0.4--10 & 1280×800 & Depth \\ \hline LiDAR Scanner & 0.5--100 & 128-beam & \makecell[c]{Point Cloud,\\ Projected Depth} \\ \hline 4D Radar System & 0.2--170 & -- & \makecell[c]{Point Cloud,\\ Projected Depth} \\ \bottomrule \end{tabular} } \vspace{-5mm} \end{table}

\vspace{-5mm}
\paragraph{RGB Camera.} As the most common visual sensor, it captures rich texture and appearance cues in the visible spectrum, supporting conventional gait representations such as silhouettes and pose sequences. Due to its ubiquity and high visual fidelity, the RGB modality serves as a natural reference for most gait recognition studies~\cite{li2024aerialgait,wang2025gaitc,xiong2024causality,wang2025ra,chen2025edinogait}.

\vspace{-5mm}
\paragraph{IR Camera.} By capturing infrared light with wavelengths greater than 780 nm, the IR camera enables reliable perception in low-light and nighttime environments. It provides near-infrared information that is invariant to illumination changes, making it widely used in RGB-IR cross-modal retrieval and recognition tasks~\cite{tan2006efficient,nahar2024infrared,manssor2021real}.

\vspace{-5mm}
\paragraph{Depth Camera.} The depth sensor employs the Time-of-Flight (ToF) principle to measure fine-grained spatial geometry, offering dense 3D body contours with minimal texture dependence~\cite{haque2016recurrent,shahroudy2016ntu}. It enhances structural understanding and bridges the gap between 2D and 3D representations.

\vspace{-5mm}
\paragraph{LiDAR Scanner.} By actively measuring the distance between the sensor and surrounding objects through laser reflections, LiDAR produces dense and highly accurate point clouds. These data provide precise 3D structural representations that are robust to lighting variations and background clutter. LiDAR-based gait data also demonstrate strong potential for identity recognition, as evidenced by its success in autonomous driving and 3D perception tasks~\cite{guo2024lidar,ahn20222v,shen2025lidargait++}.

\vspace{-5mm}
\paragraph{4D Radar System.} The Radar sensor emits frequency-modulated continuous waves (FMCW) and analyzes the reflected signals in both time and frequency domains. It supports long-range motion sensing with strong penetration capability in challenging conditions such as fog, rain, or partial occlusion. Owing to its robustness and low cost, radar-based systems have been increasingly explored for human activity recognition and through-wall monitoring~\cite{cheng2021person,hanif2022micro,meng2020gait,wang2019human,huang2025l4dr}.

In our setup, all sensors operate at a 10 Hz sampling frequency. The RGB and depth data are captured using a shared Orbbec Gemini 2 XL camera module~\cite{Orbbec}, which enables natural synchronization between these modalities. The infrared data are recorded by an industrial IR camera operating at a 940 nm narrowband wavelength to avoid interference with the LiDAR’s laser signals, ensuring clean and consistent data acquisition. The LiDAR data are collected using an Ouster OS0-7 scanner~\cite{ouster}, while the 4D radar data come from a GPAL Ares-R7861 radar system~\cite{radar}.
A summary of the sensor specifications and corresponding modalities is provided in Table~\ref{Tab:sensor}.

\subsection{Data Collection Setup}
\begin{figure}[t]
	\centering
	\begin{minipage}[t]{0.38\linewidth}
		\centering
		\includegraphics[width=\linewidth]{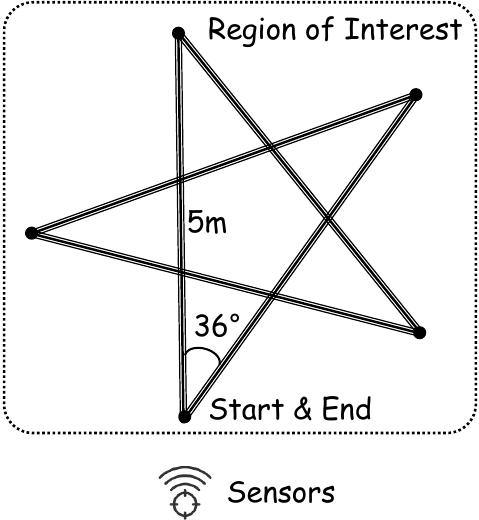}
		\caption{The collection setup of MMGait.}
		\label{Fig:collection}
	\end{minipage}
	\hfill
	\begin{minipage}[t]{0.58\linewidth}
		\centering
		\includegraphics[width=\linewidth]{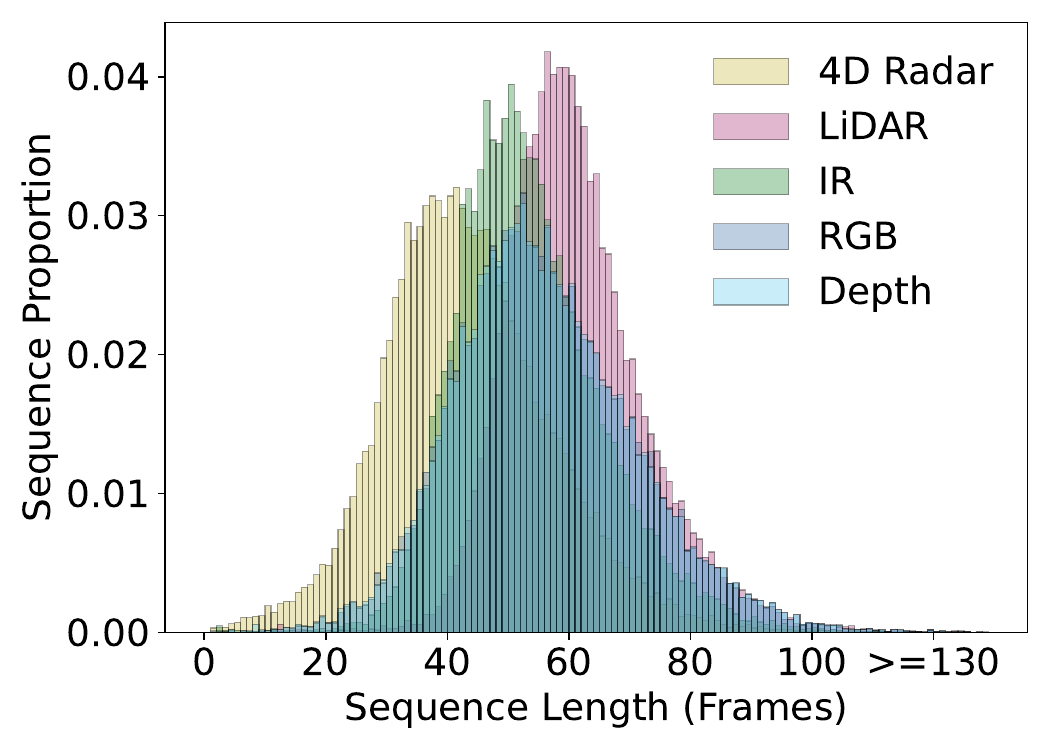}
		\caption{Sequence length distribution of MMGait.}
		\label{Fig:sta}
	\end{minipage}
	\vspace{-5mm}
\end{figure}

Our MMGait benchmark is acquired along a five-pointed star walking route, as illustrated in Figure~\ref{Fig:collection}. This route design offers full-circle coverage, capturing gait data from 10 evenly spaced viewpoints spanning $0^\circ$ to $360^\circ$ at $36^\circ$ intervals. Such a configuration ensures rich multi-view information for each walking sequence, which is essential for evaluating cross-view recognition. The recording spanned an entire month, covering different times of day from morning to evening, and including both sunny and cloudy days. This setup naturally introduces diverse lighting conditions. Participants are asked to walk the route under three distinct conditions: twice under normal walking (NM), once with a backpack (BG), and once with changed clothing (CL). Notably, our setup emphasizes real-world diversity by requiring participants to \textit{use their own backpacks} and \textit{independently change both upper and lower garments}. This enhances cross-condition realism and practical relevance.

Ideally, each subject contributes 480 sequences (10 views $\times$ (2 + 1 + 1) conditions $\times$ 12 modalities). We design customized preprocessing pipelines for each modality to ensure high data quality.  In practice, some sequences are excluded due to poor data quality or sensor failures during acquisition. After filtering, the MMGait dataset comprises data from 725 subjects, resulting in a total of 334,060 gait sequences. Detailed preprocessing pipelines and more visualization examples are provided in the supplementary materials. Figure~\ref{Fig:sta} shows the sequence length distribution of each sensor.
We summarize existing publicly available gait recognition datasets in Table~\ref{Tab:dataset}, highlighting the diverse modalities introduced in MMGait.

\begin{table}[t]
	\caption{Comparison of existing gait recognition datasets with various capture sensors.}
	\centering
	\renewcommand{\arraystretch}{1.2} 
	\label{Tab:dataset}
    \small
	\begin{adjustbox}{max width=\linewidth}
		\begin{tabular}{ccccc}
			\toprule
			Dataset   & \# Subject & \# Seq. & \# View & Sensor    \\ \midrule
            \rowcolor{blue!10}
			\multicolumn{5}{c}{\textbf{Single-Modal Gait Recognition Dataset}} \\
            CASIA-B (2006)~\cite{yu2006framework}     & 124        & 13,640  & 11      & RGB Camera                     \\
			SZTAKI-LGA (2016)~\cite{benedek2016lidar}  & 28         & 11     & 1       & LiDAR   \\
			PCG (2020)~\cite{yamada2020gait}      & 30         & 60     & 1       & LiDAR                 \\
			GREW (2021)~\cite{zhu2021gait}     &    26,345    &  128,671  &   882     & RGB Camera       \\
			Gait3D (2022)~\cite{zheng2022gait}    & 4,000       & 25,309  & 39      & RGB Camera   \\
			CCPG (2023)~\cite{li2023depth}   & 200        & 16,566  & 10      & RGB Camera               \\
			CCGR (2024)~\cite{zou2024cross}    &  970      & 1,580,617   &  33      & RGB Camera  \\ 

			\hline
            \rowcolor{blue!10}
			\multicolumn{5}{c}{\textbf{Multi-Modal Gait Recognition Dataset}} \\
			CASIA-C (2006)~\cite{tan2006efficient}   & 153      & 1,530 & 1      & RGB, IR Camera           \\
			TUM-GAID (2012)~\cite{hofmann2014tum}   & 305      & 3,370 & 1      & RGB, Depth Camera, Audio       \\
			SUSTech1K (2023)~\cite{shen2023lidargait}  & 1,050       & 25,239  & 12      & RGB Camera, LiDAR    \\ 
			FreeGait (2024)~\cite{han2024gait}  & 1,195       & 11,950  & 1     & RGB Camera, LiDAR     \\		
			\textbf{MMGait (Ours)}    & \textbf{725}        & \textbf{334,060}    &  \textbf{10}       & \textbf{\begin{tabular}[c]{@{}c@{}}RGB, Depth, IR Camera\\ LiDAR, 4D Radar\end{tabular}} \\ \hline
		\end{tabular}
	\end{adjustbox}
	\vspace{-5mm}
\end{table}

\subsection{Evaluation Protocols}

To ensure fair and consistent evaluation, MMGait is divided into a training set with 200 subjects and a test set with 525 subjects.
The training set includes diverse modalities and walking conditions to support multi-modal representation learning, while the test set contains entirely unseen identities to rigorously assess generalization.

For \textit{single-modal recognition}, both the gallery and query are constructed from the same modality, where the NM-01 sequence serves as the gallery and NM-02, BG-01, and CL-01 are used as queries.

For \textit{cross-modal recognition}, the gallery and query come from different modalities. Specifically, NM-01 sequences of one modality form the gallery, while NM-02, BG-01, and CL-01 from another modality are used as queries.

For \textit{multi-modal recognition}, the evaluation protocol follows the same gallery-query split as the single-modal setting. The only difference is that each representation is obtained by fusing features from two different modalities.

Following standard practice, evaluation metrics include Rank-1 accuracy (R1) and mean Average Precision (mAP)~\cite{fan2023opengait}. The reported results are averaged over all 10$\times$10 cross-view pairs, excluding same-view pairs.

 \section{Experiments}
\label{sec:benchmark}

Our goal is to \textbf{\textit{provide a fair and systematic exploration of diverse sensing modalities within a consistent and reproducible baseline framework}}. Accordingly, in this section, we follow established general-purpose network architectures for gait recognition and adopt representative frameworks for different experimental settings. 
For completeness, detailed model configurations, training strategies are provided in the supplementary material.
\begin{table}[t]
	\centering
	\caption{Baseline performance of silhouette- and pose-based gait recognition methods on MMGait. Results are reported in \%. Bold numbers indicate the best performance within each modality.}
	\label{tab:sil_pose}
	\renewcommand{\arraystretch}{1.05}
	\setlength{\tabcolsep}{4.3pt}
	\small
	\vspace{-2mm}
	\begin{adjustbox}{max width=\linewidth}
		\begin{tabular}{llcccccc}
			\toprule
			\multirow{2}{*}{Modality} & \multirow{2}{*}{Method} &
			\multicolumn{2}{c}{NM} & 
			\multicolumn{2}{c}{BG} & 
			\multicolumn{2}{c}{CL} \\
			\cmidrule(lr){3-4} \cmidrule(lr){5-6} \cmidrule(lr){7-8}
			& & R1 & mAP & R1 & mAP & R1 & mAP \\
			\midrule
			\multirow{5}{*}{Silhouette}
			& GaitSet~\cite{chao2019gaitset} & 95.2 & 96.8 & 90.4 & 93.4 & 48.9 & 60.7 \\
			& GaitPart~\cite{fan2020gaitpart} & 96.4 & 97.6 & 92.2 & 94.9 & 52.0 & 63.5 \\
			& GaitGL~\cite{lin2022gaitgl} & 93.1 & 95.1 & 89.2 & 92.4 & 49.5 & 60.8 \\
			& GaitBase~\cite{fan2023opengait} & 98.5 & 99.0 & 96.4 & 97.6 & \textbf{61.0} & \textbf{71.3} \\
			& DeepGaitV2-P3D~\cite{fan2023exploring} & \textbf{98.7} & \textbf{99.1} & \textbf{97.0} & \textbf{98.1} & 58.7 & 69.1 \\
			\midrule
			\multirow{6}{*}{2D Pose}
			& GaitGraph~\cite{teepe2021gaitgraph} & 31.5 & 44.8 & 23.5 & 36.0 & 13.7 & 24.4 \\
			& GaitGraph2~\cite{teepe2022towards} & 15.8 & 23.4 & 12.0 & 19.1 & 7.1 & 13.6 \\
			& GaitTR~\cite{zhang2023spatial} & 72.6 & 80.3 & 56.6 & 67.0 & 38.1 & 50.3 \\
			& GPGait~\cite{fu2023gpgait} & 80.4 & 85.7 & 66.0 & 74.0 & 40.9 & 52.1 \\
			& GPGait++~\cite{meng2025fastposegait} & \textbf{84.7} & \textbf{89.3} & \textbf{71.6} & \textbf{79.1} & \textbf{51.2} & \textbf{61.6} \\
			& SkeletonGait~\cite{fan2024skeletongait} & 82.7 & 87.5 & 71.0 & 78.2 & 43.4 & 54.2 \\
			\bottomrule
		\end{tabular}
	\end{adjustbox}
	\vspace{-5mm}
\end{table}

\subsection{Single-Modal Recognition}

This set of experiments has two main objectives. First, we evaluate representative methods on well-established modalities (\textit{e.g.}, silhouettes and 2D pose sequences) to establish performance baselines. Second, we systematically assess the standalone identification capabilities of emerging modalities in MMGait, which have received limited attention in previous gait recognition literature.

\subsubsection{Baseline Evaluation on Silhouette and Pose}
Table~\ref{tab:sil_pose} compares several state-of-the-art gait recognition methods that utilize either silhouette or 2D pose sequences. We evaluate eleven widely adopted models in total, including five silhouette-based methods and six pose-based methods. Among the silhouette-based methods, GaitBase~\cite{fan2023opengait} and DeepGaitV2-P3D (referring to the CCPG configuration)~\cite{fan2023exploring} achieve the highest accuracy, particularly under the clothing variation condition, demonstrating strong robustness. For pose-based approaches, SkeletonGait~\cite{fan2024skeletongait} and GPGait++~\cite{meng2025fastposegait} show significant improvements over earlier methods, highlighting advances in temporal modeling and structural understanding.

\subsubsection{Extending Analysis to Emerging Modalities}

We extend our analysis to a broader range of sensor-driven modalities available in MMGait. These include IR, Depth, Event, 3D Pose, LiDAR point cloud, and 4D Radar point cloud, which represent emerging directions in gait recognition research.
To ensure fair and meaningful comparisons, \textit{our experiments focus on evaluating the recognition capability of each modality rather than comparing different network architectures.} Therefore, we adopt representative and widely used baseline models and corresponding training strategies for each modality. 
For visual modalities such as RGB, Event, IR, and Projected Depth, we adopt GaitBase~\cite{fan2023opengait} as the baseline due to its simplicity and effectiveness. For 2D and 3D pose, we use GPGait++~\cite{meng2025fastposegait} due to its competitive performance. For point-based modalities (LiDAR and 4D Radar point clouds), we use LidarGait++~\cite{shen2025lidargait++}.

Table~\ref{tab:single_modal} shows that among the RGB-derived modalities, RGB images and silhouettes achieve the highest recognition performance, whereas other modalities still face certain challenges. Notably, the IR modality achieves promising accuracy under the clothing variation condition, a similar phenomenon observed in other benchmarks~\cite{li2025towards}. This advantage likely stems from its inherent ability to suppress texture and color variations while preserving discriminative structural cues. In addition, structural-oriented modalities such as Depth and LiDAR also perform well under clothing changes, whereas sparse modalities such as 4D Radar exhibit more limited performance.
For silhouette representations, the performance gap between RGB and IR silhouettes is mainly attributed to the segmentation domain shift, since the silhouette extractor is trained on RGB data and does not generalize perfectly to infrared imagery, leading to less accurate foreground masks.
\begin{table}[t]
	\centering
	\caption{Recognition performance across diverse and emerging gait sensing modalities. 
		Results are reported in \%. Bold numbers indicate the best performance within each sensor type.}
	\label{tab:single_modal}
	\renewcommand{\arraystretch}{1.05}
	\small
	\setlength{\tabcolsep}{4.3pt}
	\begin{adjustbox}{max width=\linewidth}
		\begin{tabular}{llcccccc}
			\toprule
			\multirow{2}{*}{Sensor} & \multirow{2}{*}{Modality} &
			\multicolumn{2}{c}{NM} & 
			\multicolumn{2}{c}{BG} & 
			\multicolumn{2}{c}{CL} \\
			\cmidrule(lr){3-4} \cmidrule(lr){5-6} \cmidrule(lr){7-8}
			& & R1 & mAP & R1 & mAP & R1 & mAP \\
			\midrule
			\multirow{5}{*}{RGB Camera} 
			& RGB & \textbf{99.7} & \textbf{99.8} & \textbf{99.1} & \textbf{99.4} & 60.7 & \textbf{71.5} \\
			& Silhouette & 98.5 & 99.0 & 96.4 & 97.6 & \textbf{61.0} & 71.3 \\
			& 2D Pose & 84.7 & 89.3 & 71.6 & 79.1 & 51.2 & 61.6 \\
			& 3D Pose & 42.8 & 54.3 & 30.2 & 41.6 & 22.2 & 33.0 \\
			& Event & 70.1 & 79.0 & 62.0 & 72.1 & 13.5 & 23.1 \\
			\midrule
			\multirow{2}{*}{IR Camera}
			& IR & \textbf{99.7} & \textbf{99.8} & \textbf{99.0} & \textbf{99.3} & \textbf{78.8} & \textbf{85.7} \\
			& Silhouette & 96.8 & 97.9 & 93.1 & 95.4 & 46.7 & 58.8 \\
			\midrule
			Depth Camera & Depth & 95.1 & 96.9 & 91.0 & 94.2 & 70.3 & 79.7 \\
			\midrule
			\multirow{2}{*}{LiDAR}
			& Projected Depth & 94.1 & 96.2 & 90.8 & 94.0 & 74.5 & 82.6 \\
			& Point Cloud & \textbf{97.0} & \textbf{98.2} & \textbf{94.8} & \textbf{96.9} & \textbf{76.6} & \textbf{84.7} \\
			\midrule
			\multirow{2}{*}{4D Radar}
			& Projected Depth & 22.4 & 36.3 & 17.6 & 30.7 & 16.0 & 28.8 \\
			& Point Cloud & \textbf{39.5} & \textbf{53.9} & \textbf{31.6} & \textbf{45.6} & \textbf{28.3} & \textbf{43.3} \\
			\bottomrule
		\end{tabular}
	\end{adjustbox}
	\vspace{-5mm}
\end{table}

\subsection{Cross-Modal Recognition}

To establish a fair baseline for cross-modal gait recognition, we build a two-stream recognition framework inspired by prior cross-modal gait studies~\cite{guo2024camera}. The architecture contains modal-specific early layers and shared deeper layers to balance representation specificity and alignment. The network is jointly optimized using a combination of cross-entropy loss and a cross-modal triplet loss to enhance both discriminative and modal-invariant representation learning.

We observe that the batch normalization statistics of shared layers strongly affect cross-modal retrieval performance. When features from different modalities are included in the same batch, the batch normalization layers receive mixed-domain statistics. This joint normalization promotes implicit feature-level alignment across modalities, resulting in more consistent representations and higher cross-modal retrieval accuracy~\cite{huang2023normalization,chang2019domain}.
To maintain stable feature statistics and enable a unified modeling framework, we adopt a consistent image-based representation for all sensors: RGB and IR cameras are modeled using silhouettes, the depth camera is represented with depth maps, and LiDAR and 4D Radar are encoded as projected depth maps.

\begin{table}[t]
		\centering
		\caption{RGB-Centered Cross-Modal Retrieval. Results are reported in \%. }
		\label{tab:cross_img2img}
		\small
		\renewcommand{\arraystretch}{1.05}
		\begin{adjustbox}{max width=\linewidth}
			\begin{tabular}{lcccccc}
				\toprule
				\multirow{2}{*}{Probe $\rightarrow$ Gallery} 
				& \multicolumn{2}{c}{NM} 
				& \multicolumn{2}{c}{BG} 
				& \multicolumn{2}{c}{CL} \\
				\cmidrule(lr){2-3} \cmidrule(lr){4-5} \cmidrule(lr){6-7}
				& R1 & mAP 
				& R1 & mAP 
				& R1 & mAP \\
				\midrule
				RGB (Sil.) $\rightarrow$ IR (Sil.) & 95.7 & 97.1 & 90.8 & 93.8 & 45.3 & 57.5 \\
				IR (Sil.) $\rightarrow$ RGB (Sil.) & 95.6 & 97.0 & 90.6 & 93.6 & 45.9 & 57.8 \\
				\midrule
				RGB (Sil.) $\rightarrow$ Depth & 76.1 & 83.1 & 67.3 & 76.1 & 30.1 & 43.3 \\
				Depth $\rightarrow$ RGB (Sil.) & 77.5 & 84.5 & 60.7 & 71.1 & 29.3 & 42.5 \\
				\midrule
				RGB (Sil.) $\rightarrow$ LiDAR (Proj. Depth) & 76.8 & 80.8 & 68.4 & 74.6 & 35.6 & 47.5 \\
				LiDAR (Proj. Depth) $\rightarrow$ RGB (Sil.) & 75.8 & 80.6 & 64.8 & 72.2 & 35.3 & 47.1 \\
				\midrule
				RGB (Sil.) $\rightarrow$ 4D Radar (Proj. Depth) & 3.6 & 8.9 & 2.8 & 7.6 & 2.2 & 6.7 \\
				4D Radar (Proj. Depth) $\rightarrow$ RGB (Sil.) & 2.3 & 6.8 & 1.6 & 5.5 & 1.6 & 5.4 \\
				\bottomrule
			\end{tabular}
		\end{adjustbox}
		\vspace{-4.2mm}
	\end{table}
	
	\begin{figure}[t]
		\centering
		\includegraphics[width=1.0\linewidth]{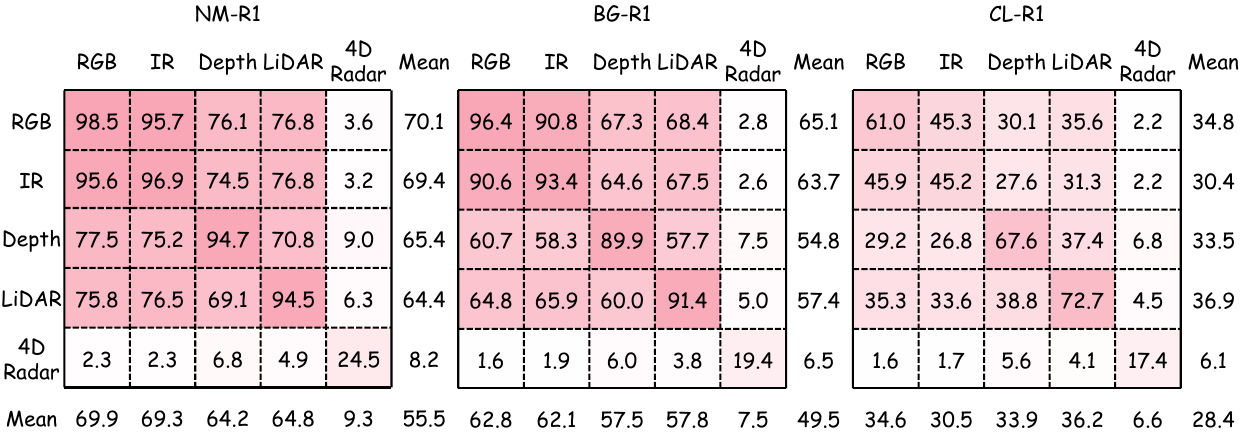}
		\vspace{-7mm}
		\caption{Pairwise cross-modal retrieval performance among the five sensors.}
		\label{Fig:cross_heatmap}
		\vspace{-7mm}
	\end{figure}

Given that RGB cameras are the most widely used and accessible visual sensors, we first focus on RGB-centered cross-modal retrieval, where RGB silhouettes serve as either the probe or the gallery. Table~\ref{tab:cross_img2img} reports retrieval performance across different modality pairs under various conditions. Retrieval between RGB and infrared silhouettes is relatively easy due to their visual similarity. In contrast, RGB$\leftrightarrow$Depth and RGB$\leftrightarrow$Projected Depth retrieval is more challenging, particularly under clothing variations, highlighting the impact of modality discrepancies. The 4D Radar modality, constrained by its inherent sparsity, shows limited suitability as a primary retrieval source and is better utilized as an auxiliary modality in multi-modal settings.

We further examine pairwise cross-modal retrieval among the five sensing modalities, as illustrated in Figure~\ref{Fig:cross_heatmap}. Ten bidirectional cross-modal models are trained, each capable of retrieving between two modalities, with diagonal entries corresponding to single-modal recognition results. The results show that cross-modal models learn relatively consistent representations across modalities, without strong asymmetry between retrieval directions. However, performance degrades substantially under more challenging conditions: while NM-R1 maintains an average accuracy of 55.5\%, BG-R1 decreases to 49.5\%, and CL-R1 further drops to 28.4\%. These observations indicate that occlusions and clothing variations significantly amplify inter-modal discrepancies, and cross-modal retrieval in scenarios with such covariate variations remains a major challenge.

\subsection{Multi-Modal Recognition}

To further investigate the benefit of leveraging complementary cues from multiple sensing modalities, we conduct multi-modal fusion experiments under both intra-sensor and inter-sensor configurations. As shown in Table~\ref{tab:intra_inter_sensor}, we follow the two-stream fusion strategy in MultiGait++~\cite{jin2025exploring}, where features from two modalities are jointly optimized while maintaining modality-specific encoders in early layers. This design enables the network to exploit both shared and complementary representations.

Under the \textbf{intra-sensor setting}, where all modalities originate from the same RGB camera, moderate gains are observed when incorporating additional cues such as event streams or human pose heatmaps. Specifically, combining RGB silhouettes with event or pose information improves performance on the challenging CL condition by +1.1\% and +3.0\% in Rank-1 accuracy, respectively. These improvements demonstrate that even within the same sensor, fusing heterogeneous signals enhances robustness to covariates.

Furthermore, the \textbf{inter-sensor setting} yields more substantial improvements, highlighting the benefit of cross-sensor complementarity. When fusing RGB silhouettes with depth or LiDAR projected depth, the model achieves a remarkable +19.3\% and +19.7\% Rank-1 improvement on CL, respectively. This indicates that depth-aware modalities provide crucial structural cues that compensate for the appearance ambiguity in RGB imagery, particularly under significant appearance changes. Notably, these gains are achieved without sacrificing performance in NM and BG conditions, suggesting effective cross-sensor feature fusion.

Beyond 2D projections, we further investigate the synergy between 3D sensing modalities. Using LiDAR point clouds as the baseline, incorporating Radar point clouds leads to an additional improvement from 76.6\% to 80.4\% in Rank-1 and from 84.7\% to 87.3\% in mAP under the CL condition. This enhancement suggests that Radar’s motion-sensitive features complement LiDAR’s spatial precision, enabling richer 3D representations and better generalization in complex environments. 

\begin{table}[t]
	\centering
	\caption{Performance comparison under Intra- and Inter-Sensor settings.
	Green numbers denote improvement over the baseline.}
	\label{tab:intra_inter_sensor}
	\small
	\renewcommand{\arraystretch}{1.1}
	\setlength{\tabcolsep}{3pt}
	\begin{adjustbox}{max width=\linewidth}

		\begin{tabular}{lcccccc}
			\toprule
			\multirow{2}{*}{Modality} & 
			\multicolumn{2}{c}{NM} & 
			\multicolumn{2}{c}{BG} & 
			\multicolumn{2}{c}{CL} \\
			\cmidrule(lr){2-3} \cmidrule(lr){4-5} \cmidrule(lr){6-7}
			& R1 & mAP & R1 & mAP & R1 & mAP \\
			\midrule
			\rowcolor{blue!10}
			\multicolumn{7}{c}{\textbf{Intra-Sensor}} \\
			RGB(Sil.) & 98.5 & 99.0 & 96.1 & 97.6 & 61.0 & 71.3 \\
			+RGB(Event) & 98.9 & 99.1 & 98.1 & 98.6 & 
			62.1 {\tiny \textcolor{green!60!black}{(+1.1)}} & 
			72.3 {\tiny \textcolor{green!60!black}{(+2.0)}} \\
			+RGB(Pose) & 98.6 & 99.0 & 97.1 & 98.0 & 
			64.0 {\tiny \textcolor{green!60!black}{(+3.0)}} & 
			73.5 {\tiny \textcolor{green!60!black}{(+2.2)}} \\
			\midrule
			\rowcolor{blue!10}
			\multicolumn{7}{c}{\textbf{Inter-Sensor}} \\
			RGB(Sil.) & 98.5 & 99.0 & 96.1 & 97.6 & 61.0 & 71.3 \\
			+Depth & 99.5 & 99.7 & 98.8 & 99.2 & 
			80.3 {\tiny \textcolor{green!60!black}{(+19.3)}} & 
			86.7 {\tiny \textcolor{green!60!black}{(+15.4)}} \\ 
			+LiDAR(Proj. Depth) & 99.7 & 99.8 & 98.8 & 99.3 & 
			80.7 {\tiny \textcolor{green!60!black}{(+19.7)}} & 
			87.1 {\tiny \textcolor{green!60!black}{(+15.8)}} \\ 
			\midrule
			LiDAR(Point Clouds) & 97.0 & 98.2 & 94.8 & 96.9 & 
			76.6  & 
			84.7  \\
			+Radar(Point Clouds) & 96.3 & 97.9 & 93.9 & 96.3 & 80.4{\tiny \textcolor{green!60!black}{(+3.8)}} & 87.3{\tiny \textcolor{green!60!black}{(+2.6)}} \\
			\bottomrule
		\end{tabular}
	\end{adjustbox}
	\vspace{-5mm}
\end{table}
 \section{Omni Multi-Modal Gait Recognition}

We have systematically evaluated three representative scenarios in gait recognition, namely s\textit{ingle-modal recognition, cross-modal recognition,} and \textit{multi-modal recognition}. Although these three paradigms have achieved remarkable progress, existing studies usually treat them as independent research tracks, and tend to focus on a restricted set of modality combinations. In contrast, real-world gait recognition systems often face much more complex and flexible query demands~\cite{ye2025transformer}. For instance, users may need to identify a person based on arbitrary sensor inputs, or perform retrieval across different modality combinations. Existing single-purpose models cannot fully support such flexibility.

To address these limitations, we propose the \textbf{Omni Multi-Modal Gait Recognition (OMGR)} task, which aims to develop a unified model capable of accepting any modality as a query and retrieving targets from any modality. The goal of OMGR is to create a single framework with three core capabilities. First, \textbf{Universal Modality Recognition} ensures reliable identification from any individual sensing modality, enabling strong generalization across heterogeneous inputs. Second, \textbf{Adaptive Multi-Modal Fusion} allows the model to exploit complementary information from specific pairs of modalities when available, improving robustness and accuracy. Third, \textbf{Modality-Agnostic Retrieval} supports flexible, bidirectional matching between arbitrary modality pairs, facilitating seamless cross-sensor retrieval.

By integrating heterogeneous modalities into a single model, OMGR moves toward a general-purpose, adaptive, and lightweight gait recognition paradigm, aligning with practical requirements for flexible human identification in real-world scenarios.

\subsection{OmniGait Framework}

Based on the previous analysis of single-purpose models and the challenges in omni multi-modal gait recognition, we present a simple yet strong baseline model, \textbf{OmniGait}. OmniGait is designed to flexibly handle single-modal, cross-modal, and multi-modal gait recognition within a unified architecture. 

Specifically, to enable unified modeling across heterogeneous inputs, OmniGait considers nine image-based modalities captured from five sensing devices, including RGB, RGB silhouette, 2D pose (represented as heatmaps~\cite{fan2024skeletongait}), event, infrared (IR), IR silhouette, depth, LiDAR-projected depth, and radar-projected depth. OmniGait supports both single-modal retrieval within each modality and arbitrary cross-modal retrieval between any two modalities. 
For multi-modal fusion experiments, we take the RGB silhouette as the anchor modality and fuse it individually with each of the remaining modalities, since RGB silhouette is the most widely adopted representation in gait recognition.

\begin{figure}[t]
	
	\centering
	\includegraphics[width=\linewidth]{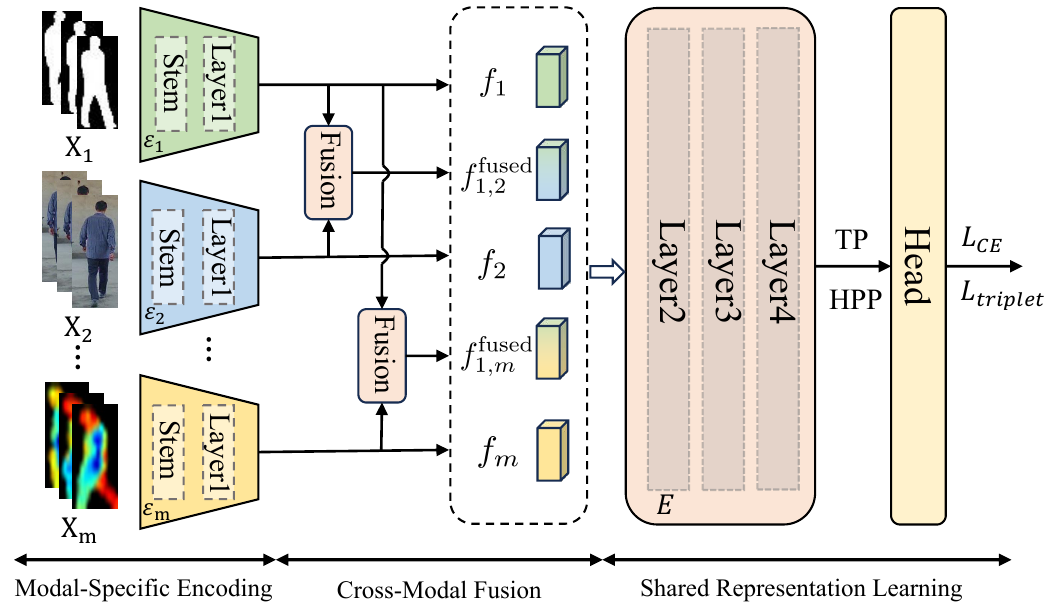}
	\caption{Overview of the OmniGait framework, including modal-specific encoding, adaptive cross-modal fusion, and shared representation learning to flexibly handle single-modal, cross-modal, and multi-modal gait recognition.}
	\label{Fig:framework}
	\vspace{-5mm}
\end{figure}

As shown in Figure~\ref{Fig:framework}, OmniGait consists of three components: (i) a modal-specific encoding stage that independently extracts discriminative representations from each modality while preserving their intrinsic physical characteristics; (ii) a cross-modal fusion module that facilitates efficient information exchange and complementary feature aggregation across heterogeneous modality pairs; and (iii) a shared representation learning mechanism that aligns unimodal and multimodal features within a unified embedding space to support diverse recognition tasks.

\vspace{-5mm}
\paragraph{Modal-Specific Encoding.}
Different sensing modalities (e.g., RGB, infrared, depth, and event) capture distinct physical cues of human motion.  
For each modality $m \in \{1,2,\dots,M\}$, an independent encoder $\varepsilon_m(\cdot)$ maps the raw input $\mathbf{X}_m$ into a feature map:
\begin{equation}
	f_m = \varepsilon_m(\mathbf{X}_m), \quad f_m \in \mathbb{R}^{T\times  C_1\times H_1\times W_1},
\end{equation}
where $T$ is the frame number, $C_1$ is the channel number, $H_1,W_1$ is the shape of the feature map. Each encoder preserves modality-specific characteristics before feature alignment or fusion.

\vspace{-5mm}
\paragraph{Cross-Modal Fusion.} To effectively integrate complementary cues from multiple modalities, we design a lightweight Cross-Modal Fusion module that adaptively balances modality contributions through gated weighting.  
Given two modality-specific features $f_i, f_j $, we first concatenate them along the channel dimension and apply a $1\times1$ convolution layer to obtain a fused representation:
\begin{equation}
	f_{i,j} = \text{Conv}_{1\times1}(\text{Concat}([f_i, f_j],\text{dim}=1))
\end{equation}

To capture the relative importance of each modality across different spatial contexts, we employ a gating mechanism $G(\cdot)$ that generates adaptive weights:
\begin{equation}
	\mathbf{w} = \text{Softmax}(G(f_{i,j})), \quad 
	\mathbf{w} \in \mathbb{R}^{2}
\end{equation}
where $G(\cdot)$ consists of global average pooling followed by two $1\times1$ convolutions and non-linear activations, producing modality-specific responses.
The final fused feature is obtained by combining the gated modality features with a residual connection:
\begin{equation}
	f_{i,j}^{\text{fused}} = f_{i,j} + \sum_{m=1}^{2} \mathbf{w}_m f_m,
	f_{i,j}^{\text{fused}} \in \mathbb{R}^{T\times  C_1\times H_1\times W_1}
\end{equation}

The fusion module is shared across all modality combinations. This design allows the network to dynamically recalibrate the contribution of each modality while preserving their shared spatial semantics.

\vspace{-5mm}
\paragraph{Shared Representation Learning.}
After obtaining unimodal features $\{f_m\}_{m=1}^{M}$ and fused bimodal features $\{f_{i,j}^{\text{fused}}\}_{i\neq j}$, we jointly forward them through a shared residual backbone $E(\cdot)$ to learn modality-invariant representations:
\begin{equation}
	\mathbf{F} = E(\{f_m, f_{i,j}^{\text{fused}}\}),
	\quad \mathbf{F} \in \mathbb{R}^{T\times C_2\times H_2\times W_2}.
\end{equation}
All unimodal and fused features are processed together within the same batch, enabling the batch normalization layers inside $E(\cdot)$ to learn more robust and generalized statistics across modalities.

The resulting feature maps are aggregated via Temporal Pooling (TP) and Horizontal Pyramid Pooling (HPP), followed by a fully connected (FC) layer and BNNeck layer, yielding the final inference feature $\in \mathbb{R}^{C_3 \times P}$, where $P$ is the number of HPP partitions.

OmniGait is optimized using a joint objective that combines identity classification loss $L_{CE}$ and triplet loss $L_{triplet}$,
which simultaneously enhances feature discriminability and cross-modal alignment for omni multi-modal gait recognition.

\vspace{-4mm}
\paragraph{Testing Pipeline.}
During training, all modalities are \textit{jointly used}, while during inference only the required modality-specific encoder(s) $\varepsilon_m$ are activated.
The shared encoder $E$ supports features from arbitrary modalities, making OmniGait fully compatible with single-, cross-, and multi-modal tasks without additional overhead. 
For \textbf{single-modal} and \textbf{cross-modal} recognition, the input modality is sequentially forwarded through the \textit{Modal-Specific Encoder}, the \textit{Shared Encoder}, and subsequent aggregation modules to obtain the final embedding.
For \textbf{multi-modal fusion} scenarios, two modalities are first processed by their respective \textit{Modal-Specific Encoders}, then jointly passed through the \textit{Cross-Modal Fusion Module}, and the fused features are subsequently fed into the \textit{Shared Backbone} and following modules for representation extraction.

\subsection{OmniGait Results}

\paragraph{Experimental Setting.} 
In our experiments, the sequence length $T$ is set to 16, and the spatial resolution of all modalities is fixed at $64 \times 64$. 
After the modal-specific encoding stage, the spatial feature maps have dimensions of $C_1 \times H_1 \times W_1 = 128 \times 64 \times 64$; 
after the shared backbone, the spatial feature dimensions become $C_2 \times H_2 \times W_2 = 512 \times 16 \times 16$. 
The final inference feature has dimensions $C_3\times P = 256\times 16$.
During training, the batch size is set to $8 \times 4$, where each batch contains $8$ identities and each identity contributes $4$ sequences. 
All training strategies and optimization settings follow those in the baseline experiments to ensure fair comparisons.

\vspace{-5mm}
\paragraph{Single-Modal Recognition.}
We first validate the effectiveness of OmniGait on each individual modality.
As shown in Table~\ref{tab:omni_single}, our unified model achieves competitive performance across all nine modalities, delivering results comparable to those of modality-specific expert models reported in Table~\ref{tab:single_modal}.
However, under the more challenging CL setting, most modalities still exhibit a noticeable gap compared with their modality-specific counterparts.

\begin{table}[t]
	\centering
	\small
	\renewcommand{\arraystretch}{1.0}
	\setlength{\tabcolsep}{2.8pt}
	\caption{Single-Modal Recognition performance of OmniGait across different modalities. Results are reported in \%. Bold numbers indicate the best performance within each sensor type.}
	\label{tab:omni_single}
	\begin{adjustbox}{max width=\linewidth}
		\begin{tabular}{llcccccc}
			\toprule
			\multirow{2}{*}{Sensor} & \multirow{2}{*}{Modality} &
			\multicolumn{2}{c}{NM} & 
			\multicolumn{2}{c}{BG} & 
			\multicolumn{2}{c}{CL} \\
			\cmidrule(lr){3-4} \cmidrule(lr){5-6} \cmidrule(lr){7-8}
			& & R1 & mAP & R1 & mAP & R1 & mAP \\
			\midrule
			\multirow{4}{*}{RGB Camera} 
			& RGB & \textbf{99.7} & \textbf{99.8} & \textbf{99.0} & \textbf{99.3} & \textbf{58.5} &  \textbf{69.9} \\
			& Silhouette & 98.1 & 98.7 & 95.7 & 97.2 & 44.9 & 57.6 \\
			& 2D Pose & 73.8 & 80.8 & 58.8 & 68.5 & 26.8 & 38.0 \\
			& Event & 87.3 & 91.5 & 78.7 & 85.0 & 15.2 & 24.9 \\
			\midrule
			\multirow{2}{*}{IR Camera}
			& IR & \textbf{99.4} & \textbf{99.6} & \textbf{97.6} & \textbf{98.4} & \textbf{59.2} & \textbf{70.7} \\
			& Silhouette & 95.4 & 96.9 & 90.7 & 93.7 & 32.4 & 45.1 \\
			\midrule
			Depth Camera & Depth & 90.1 & 93.2 & 80.5 & 86.2 & 36.8 & 49.5 \\
			\midrule
			LiDAR
			& Projected Depth & 93.1 & 95.5 & 88.6 & 92.4 & 43.3 & 57.0 \\
			\midrule
			4D Radar
			& Projected Depth & 15.8 & 27.7 & 12.7 & 23.8 & 11.5 & 22.6  \\
			\bottomrule
		\end{tabular}
	\end{adjustbox}
	\vspace{-2mm}
\end{table}

\vspace{-5mm}
\paragraph{Cross-Modal Recognition.}

\begin{figure}[t]
	\centering
	\includegraphics[width=\linewidth]{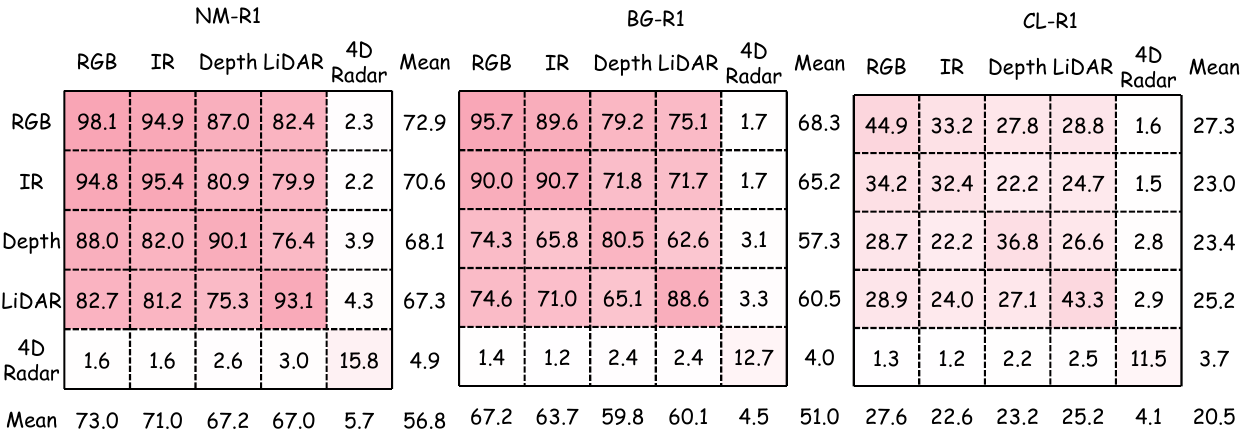}
	\caption{Pairwise cross-modal retrieval performance of OmniGait among the five sensors.}
	\vspace{-5mm}
	\label{Fig:omni_heatmap}
\end{figure}

We further investigate OmniGait’s effectiveness in learning modality-invariant representations for cross-modal retrieval.
As illustrated in Figure~\ref{Fig:omni_heatmap}, OmniGait achieves promising cross-modal retrieval performance, outperforming modality-specific cross-retrieval baselines in the NM and BG settings.
For instance, in the NM scenario, the RGB$\rightarrow$Depth retrieval achieves 87.0\% Rank-1 accuracy, notably higher than the 77.5\% of a dedicated cross-modal model trained solely on this pair.
This improvement mainly stems from the shared backbone, which enables OmniGait to learn more modality-robust statistical representations and thereby provides an effective bridge for aligning heterogeneous features.
Nonetheless, performance under the CL condition remains relatively low, indicating extreme clothing variations still disrupt the fine-grained appearance correspondence between modalities.

\vspace{-5mm}
\paragraph{Multi-Modal Recognition.}
Finally, we examine the fusion performance under both intra-sensor and inter-sensor settings, as summarized in Table~\ref{tab:omni_intra_inter_sensor}.
Using RGB silhouette as the anchor modality, the integration of complementary signals consistently improves recognition, particularly in the challenging CL setting.
For instance, fusing RGB image with RGB silhouette yields a substantial boost of +19.8\% in Rank-1 accuracy and +17.3\% in mAP.
Similarly, under the inter-sensor configuration, fusing with depth and LiDAR-projected depth brings gains of +4.2\% and +7.0\% Rank-1, respectively.
These results confirm that OmniGait’s modular design effectively captures complementary cues from multiple sensing sources, enabling robust gait understanding under diverse conditions.

\begin{table}[t]
	\centering
	\small
	\renewcommand{\arraystretch}{1.1}
	\setlength{\tabcolsep}{3pt}
	\caption{OmniGait performance under intra- and inter-sensor settings.
	Green numbers denote improvement over the baseline.}
	\label{tab:omni_intra_inter_sensor}
	\begin{adjustbox}{max width=\linewidth}

		\begin{tabular}{lcccccc}
			\toprule
			\multirow{2}{*}{Modality} & 
			\multicolumn{2}{c}{NM} & 
			\multicolumn{2}{c}{BG} & 
			\multicolumn{2}{c}{CL} \\
			\cmidrule(lr){2-3} \cmidrule(lr){4-5} \cmidrule(lr){6-7}
			& R1 & mAP & R1 & mAP & R1 & mAP \\
			\midrule
			\rowcolor{blue!10}
			\multicolumn{7}{c}{\textbf{Intra-Sensor}} \\
			RGB(Sil.) & 98.1 & 98.7 & 95.7 & 97.2 & 44.9 & 57.6 \\
			+RGB(Event) & 98.6 & 99.1 & 96.8 & 97.9 & 
			45.4 {\tiny \textcolor{green!60!black}{(+0.5)}} & 
			58.1 {\tiny \textcolor{green!60!black}{(+0.5)}} \\
			+RGB(Pose) & 98.0 &  98.7 & 95.8 & 97.2 & 
			46.1 {\tiny \textcolor{green!60!black}{(+1.2)}} & 
			58.6 {\tiny \textcolor{green!60!black}{(+1.0)}} \\
			+RGB(RGB) & 99.8 &  99.9 & 99.5 & 99.7 & 
			64.7 {\tiny \textcolor{green!60!black}{(+19.8)}} & 
			74.9 {\tiny \textcolor{green!60!black}{(+17.3)}} \\
			\midrule
			\rowcolor{blue!10}
			\multicolumn{7}{c}{\textbf{Inter-Sensor}} \\
			RGB(Sil.) & 98.1 & 98.7 & 95.7 & 97.2 & 44.9 & 57.6 \\
			+Depth & 98.3 & 98.9 & 96.4 & 97.6 & 
			49.1 {\tiny \textcolor{green!60!black}{(+4.2)}} & 
			61.5 {\tiny \textcolor{green!60!black}{(+3.9)}} \\ 
			+Radar(Proj. Depth) & 98.4 & 98.9 & 96.4 & 97.7 & 
			50.9 {\tiny \textcolor{green!60!black}{(+6.0)}} & 
			63.1 {\tiny \textcolor{green!60!black}{(+5.5)}} \\ 
			+LiDAR(Proj. Depth) & 98.5 & 99.1 & 96.7 & 97.9 & 
			51.9 {\tiny \textcolor{green!60!black}{(+7.0)}} & 
			63.9 {\tiny \textcolor{green!60!black}{(+6.3)}} \\ 
			\bottomrule
		\end{tabular}
	\end{adjustbox}
	\vspace{-5mm}
\end{table}
 
\section{Conclusion}
In this work, we introduce MMGait, a comprehensive multi-modal gait recognition dataset, encompassing 725 subjects and 334,060 sequences across 12 modalities from 5 heterogeneous sensors.
Based on MMGait, we conduct extensive evaluations in single-modal, cross-modal, and multi-modal settings to provide a holistic understanding of modality characteristics, robustness, and complementarity. 
Furthermore, we propose the Omni Multi-Modal Gait Recognition task and present the baseline model OmniGait, validating the feasibility of unifying diverse recognition paradigms within a single framework.
Together, MMGait and OmniGait establish a solid foundation for universal gait understanding and open new directions toward scalable, flexible, and deployable gait recognition systems.

\section{Acknowledgment}
This work is jointly supported by Joint Fund for the Provincial Science and Technology R\&D Program of Henan Province (245200810009), National Natural Science Foundation of China (62476027, 62276025) and the Fundamental Research Funds for the Central Universities (2253200026).

{
    \small
    \bibliographystyle{ieeenat_fullname}
    \bibliography{main}
}
\maketitlesupplementary
\setcounter{page}{1}
\setcounter{section}{5}
\setcounter{table}{8}
\setcounter{figure}{6}
\setcounter{equation}{5}

\section{Related Work}
\subsection{Gait Recognition Methods}
Gait recognition methods are broadly classified into two categories: silhouette-based methods~\cite{zheng2024takes,peng2024glgait,dou2024clash,huang2025learning,huangvocabulary,gaitasset,hou2025gaitsnippet} and pose-based methods~\cite{guo2023physics,fu2024cut,huang2023condition}.
Silhouette-based methods rely on binary gait silhouettes, which are widely used due to their effectiveness in capturing spatio-temporal information. For example, GaitSet~\cite{chao2019gaitset} models gait as unordered frame sets using set pooling. More recent works like GaitBase~\cite{fan2023opengait} and DeepGaitV2~\cite{fan2023exploring} revisit architectural design, introducing efficient backbone networks tailored for gait recognition. Additionally, some studies~\cite{ma2024learning,ma2023dynamic} emphasize modeling specific motion patterns to enhance the temporal discriminability of gait features. 
Pose-based methods, on the other hand, utilize structural representations of the human body based on joint coordinates. PoseGait~\cite{liao2020model} leverages 3D human poses as input to CNN-based architectures to extract discriminative features. GaitGraph~\cite{teepe2021gaitgraph}, GaitGraph2~\cite{teepe2022towards}, and GPGait~\cite{fu2023gpgait} adopt graph convolutional networks (GCNs) to explicitly model spatial dependencies between joints. Furthermore, methods like SkeletonGait~\cite{fan2024skeletongait} and GaitHeat~\cite{fu2024cut} represent joint positions through heatmaps, preserving fine-grained spatial details and enabling a more comprehensive encoding of body shape and motion. 
In addition, several emerging modalities have recently been explored for gait recognition. BigGait~\cite{ye2024biggait} and BiggerGait~\cite{ye2025biggergait} leverage large vision models to extract powerful visual representations from RGB inputs. EdinoGait~\cite{chen2025edinogait} exploits the capability of large vision models to address event-based gait recognition, demonstrating clear advantages under low-light conditions. LidarGait++~\cite{shen2025lidargait++} further introduces effective local representation techniques for point-cloud-based gait recognition.

More recently, multi-modal gait recognition has gained increasing attention by integrating complementary modalities~\cite{wang2023combining}. MMGaitFormer~\cite{cui2023multi} fuses silhouettes and 2D poses through spatial–temporal modules to enhance cross-source alignment, while MultiGait++~\cite{jin2025exploring} provides a simple yet strong baseline and systematically compares diverse fusion strategies. TriGait~\cite{sun2024trigait} introduces a tri-branch hybrid fusion framework to jointly exploit the complementary cues of silhouettes and poses. LiCAF~\cite{deng2024licaf} further proposes an effective LiDAR-camera fusion scheme to obtain robust cross-sensor gait representations.
\subsection{Gait Recognition Benchmark}

Existing gait recognition benchmarks predominantly rely on silhouette or pose modalities, exemplified by indoor datasets such as CASIA-B~\cite{yu2006framework}, OU-MVLP~\cite{takemura2018multi}, CCPG~\cite{li2023depth}, and CCGR~\cite{zou2024cross}, as well as in-the-wild datasets like Gait3D~\cite{zheng2022gait} and GREW~\cite{zhu2021gait}. While effective in controlled settings, these modality choices limit scalability to diverse sensing environments and raise potential privacy concerns.
Recent advances in hardware have broadened the sensing landscape, with LiDAR emerging as a promising modality due to its resilience to lighting variations, background clutter, and occlusions. Representative efforts such as LidarGait~\cite{shen2023lidargait}, which introduces the SUSTech1K dataset, and FreeGait~\cite{han2024gait}, which extends LiDAR-based recognition to unconstrained outdoor conditions, highlight the potential of structural 3D cues for robust gait analysis.
However, these efforts still explore only a small portion of the sensing modalities used in real-world systems, leading to fragmented progress across isolated modal combinations and lacking a unified basis for comparison. A truly comprehensive multi-modal benchmark remains absent, underscoring the need for a unified and diverse dataset to advance gait recognition in realistic settings.
\subsection{Unified Models for Identity Recognition}

Recent efforts in person re-identification increasingly focus on building unified models capable of handling diverse modalities and tasks. 
Instruct-ReID++~\cite{he2025instruct} takes a significant step toward universal identity retrieval by employing natural-language instructions to guide a single model across varied scenarios, enabling task-adaptive behavior through instruction tuning and adaptive losses. 
Complementary to this task-unified perspective, the All-in-One (AIO) framework~\cite{li2024all} adopts a frozen pre-trained backbone with modality-specific designs to extract consistent identity features from RGB, infrared, sketch, and text inputs. 
ReID5o~\cite{zuo2025reid5o} further expands modality unification by introducing a five-modality benchmark and a unified encoder with multi-expert routing, achieving flexible cross-modal retrieval across arbitrary modality pairs. 
Our work aligns with this direction but focuses specifically on unifying diverse sensing modalities within a single framework.

\section{More Details in MMGait}

\subsection{Data Process Pipeline}

We constructed separate processing pipelines for each sensor, ultimately producing 12 distinct modalities for evaluation. To the best of our knowledge, MMGait is the first large-scale gait recognition dataset that covers such a comprehensive range of modalities. The details of the processing pipelines are summarized below:

\vspace{-5mm}
\paragraph{RGB Camera:}We use a pedestrian tracking network~\cite{zhang2022bytetrack} to extract bounding boxes. Instance segmentation is applied to obtain silhouettes, followed by pose estimation to extract both 2D and 3D pose representations~\cite{liu2021paddleseg,wang2020deep,cai2024disentangled,cai2025fastddhpose}. Additionally, we employ the V2E~\cite{hu2021v2e} algorithm to convert RGB videos into event-based representations. Cropped event sequences are then generated by scaling the RGB-based bounding boxes according to the resolution ratio between the event and RGB videos.

\vspace{-5mm}
\paragraph{Depth Camera:}As pedestrian tracking is challenging directly on depth data, and the RGB and depth cameras are integrated in a shared device, we computed the spatial offset between the RGB and depth cameras to adjust the bounding box coordinates obtained from the RGB videos, and used the transformed boxes to crop the corresponding depth-based gait sequences.

\vspace{-5mm}
\paragraph{IR Camera:}Pedestrian tracking and instance segmentation are performed independently on infrared videos. Both the cropped IR images and the corresponding silhouettes demonstrate high visual quality.

\vspace{-5mm}
\paragraph{LiDAR Scanner:}The raw LiDAR scans are processed into clean human walking point clouds and their corresponding projection maps.
    Specifically, we first retain points within a predefined Region of Interest (ROI) that captures the walking subject. Ground removal~\cite{lee2022patchwork++} is then performed to remove floor points, followed by denoising algorithms~\cite{ester1996density} to filter out scattered noise and irrelevant objects. The projection of point clouds into 2D depth maps is implemented based on the OpenGait  codebase~\cite{fan2023opengait,shen2023lidargait}.

\vspace{-5mm}
\paragraph{4D Radar System:}As the radar sensor captures mainly moving targets and there are no additional dynamic objects indoors, the point clouds are restricted to those falling within the specified ROI. The projection of point clouds into 2D depth maps is also implemented based on the OpenGait codebase~\cite{fan2023opengait,shen2023lidargait}.

\subsection{Visualization}

\begin{figure*}[htbp]
    \centering
    \includegraphics[width=1.1\linewidth]{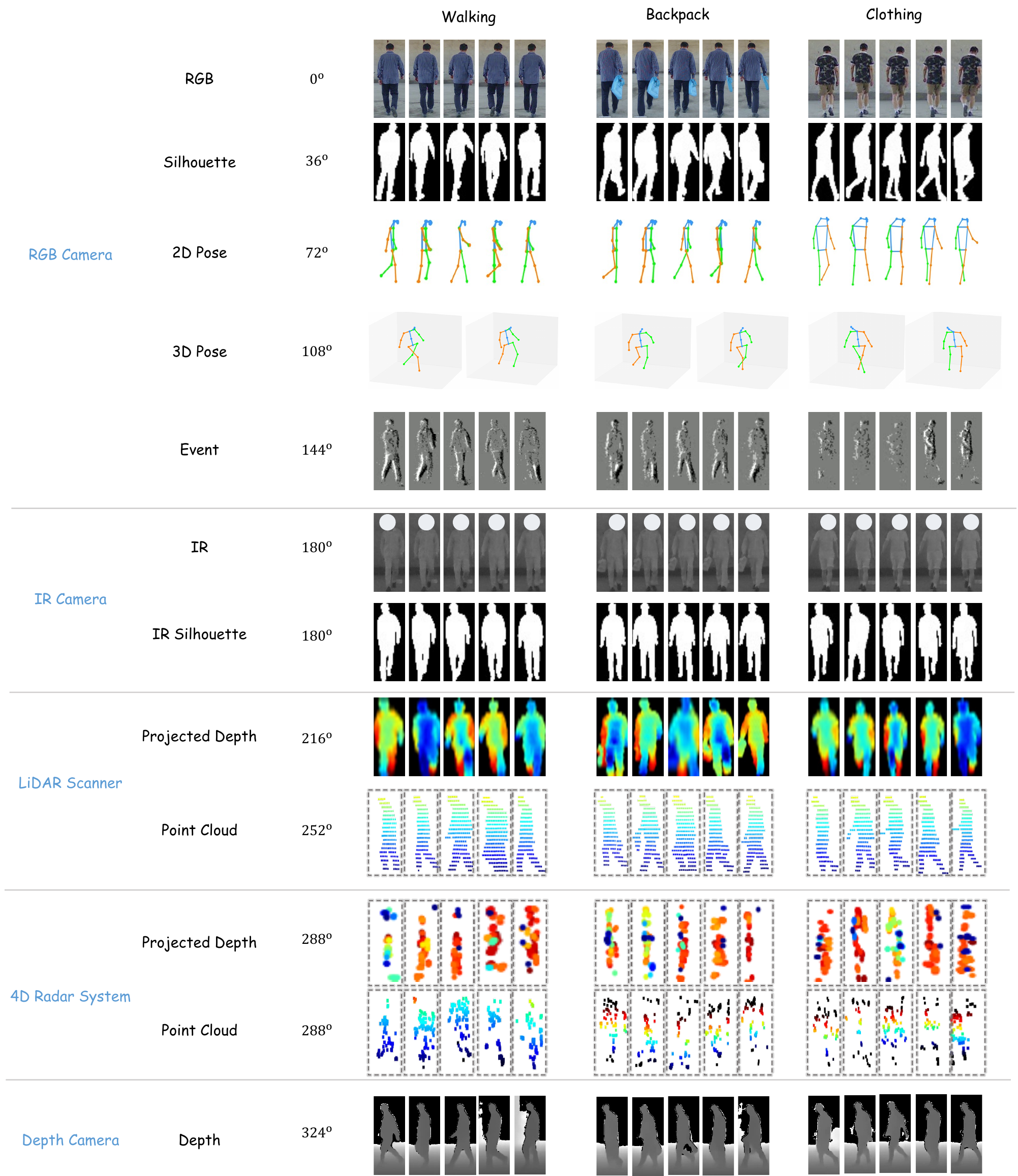} 
    \caption{Visualization of gait sequences across all sensor modalities, ten viewpoints, and three walking conditions (NM, BG, CL).}
    \label{fig:vis1}
\end{figure*}

\begin{figure*}[htbp]
    \centering
    \includegraphics[width=1.1\linewidth]{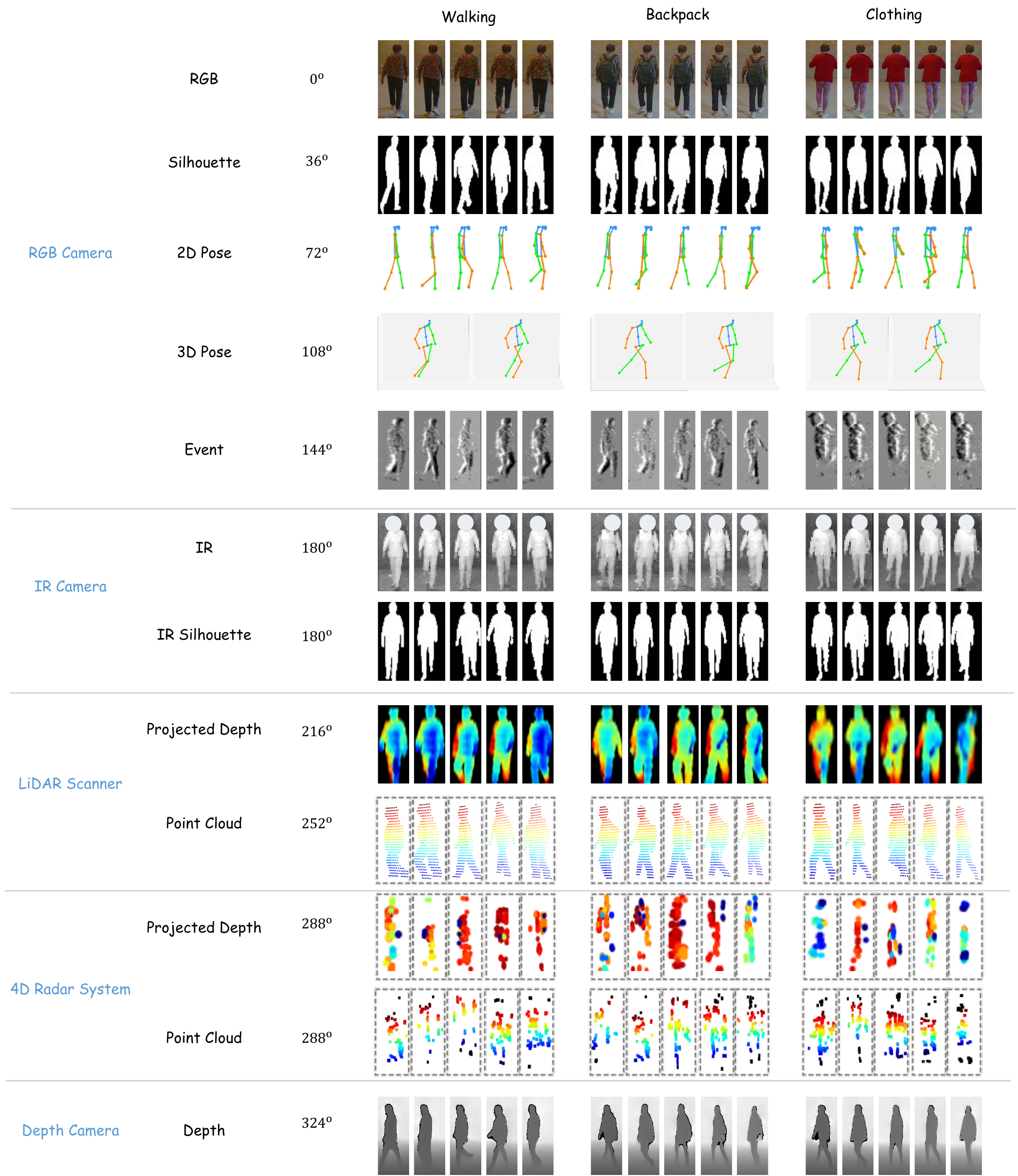} 
    \caption{Visualization of gait sequences across all sensor modalities, ten viewpoints, and three walking conditions (NM, BG, CL).}
    \label{fig:vis2}
\end{figure*}

We present supplementary visualization examples, as shown in Figure~\ref{fig:vis1} and Figure~\ref{fig:vis2}, encompassing all sensor modalities, ten viewpoints, and the three walking conditions: normal walking (NM), walking with a backpack (BG), and walking with changed clothing (CL). These samples illustrate the diversity and high quality of the collected gait sequences across different sensing configurations and walking scenarios.

\vspace{-5.5mm}

\paragraph{Privacy Statement.}
All data collection procedures followed ethical guidelines, and all participants provided written informed consent for research purposes. The dataset is released strictly for academic research, and any form of misuse or unauthorized application is explicitly prohibited.

\section{Experimental Setup}

\begin{table*}[t]
    \centering
    \small
    \begin{tabular}{l l c c}
    \toprule
    Task Setting & Modality & Parameters (M) & FLOPs (G) \\
    \midrule
    \multirow{9}{*}{\makecell[l]{Single-Modal\\(GaitBase)}}
    & RGB (Sil.) & 7.821504 & 51.67659827 \\
    & RGB (RGB) & 7.822656 & 51.81815603 \\
    & RGB (2D Pose) & 7.822080 & 51.74737715 \\
    & RGB (Event) & 7.822656 & 51.81815603 \\
    & IR (Sil.) & 7.821504 & 51.67659827 \\
    & IR (IR) & 7.822656 & 51.81815603 \\
    & Depth & 7.822656 & 51.81815603 \\
    & LiDAR (Projected Depth) & 7.822656 & 51.81815603 \\
    & 4D Radar (Projected Depth) & 7.822656 & 51.81815603 \\
    \midrule
    \multirow{10}{*}{\makecell[l]{Cross-Modal\\(Two-Stream)}}
    & RGB (Sil.) $\leftrightarrow$ IR (Sil.) & 10.820736 & 103.3531965 \\
    & RGB (Sil.) $\leftrightarrow$ Depth & 10.821888 & 103.4947543 \\
    & RGB (Sil.) $\leftrightarrow$ LiDAR (Projected Depth) & 10.821888 & 103.4947543 \\
    & RGB (Sil.) $\leftrightarrow$ 4D Radar (Projected Depth) & 10.821888 & 103.4947543 \\
    & IR (Sil.) $\leftrightarrow$ Depth & 10.821888 & 103.4947543 \\
    & IR (Sil.) $\leftrightarrow$ LiDAR (Projected Depth) & 10.821888 & 103.4947543 \\
    & IR (Sil.) $\leftrightarrow$ 4D Radar (Projected Depth) & 10.821888 & 103.4947543 \\
    & Depth $\leftrightarrow$ LiDAR (Projected Depth) & 10.823040 & 103.6363121 \\
    & Depth $\leftrightarrow$ 4D Radar (Projected Depth) & 10.823040 & 103.6363121 \\
    & LiDAR (Projected Depth) $\leftrightarrow$ 4D Radar (Projected Depth) & 10.823040 & 103.6363121 \\
    \midrule
    \multirow{4}{*}{\makecell[l]{Multi-Modal\\(MultiGait++)}}
    & RGB (Sil.) + RGB (Event) & 11.920448 & 106.2864364 \\
    & RGB (Sil.) + RGB (Pose) & 11.919872 & 106.2156575 \\
    & RGB (Sil.) + Depth & 11.920448 & 106.2864364 \\
    & RGB (Sil.) + LiDAR (Projected Depth) & 11.920448 & 106.2864364 \\
    \midrule
    \makecell[l]{Omni Multi-Modal \\ (OmniGait)}
    & \makecell[l]{RGB (Sil., RGB, 2D Pose, Event), IR (Sil., IR), Depth,\\
                   LiDAR (Projected Depth), 4D Radar (Projected Depth)}
    & 9.963266 & 1054.108566 \\
    \bottomrule
    
    \end{tabular}
    \caption{Parameters and FLOPs for different models evaluated in our experiments.}
    \label{tab:params_flops}
    \end{table*}

To comprehensively evaluate the effectiveness of \textbf{MMGait}, we design three categories of experiments: \textit{Single-Modal Recognition}, \textit{Cross-Modal Recognition}, and \textit{Multi-Modal Recognition}. These experiments aim to systematically assess the recognition capability of each modality, cross-modal retrieval performance, and the complementary relationships among different modalities under a unified and fair evaluation protocol.

All input visual modalities are standardized to a spatial resolution of $64 \times 64$. Unless otherwise stated, all configurations follow the default settings of OpenGait~\cite{fan2023opengait} and FastPoseGait~\cite{meng2023fastposegait}. For visual and pose modalities, we adopt a batch configuration of $(p, k, l) = (8, 8, 30)$, where $p$ denotes the number of identities, $k$ the number of sequences per identity, and $l$ the number of frames. For LiDAR and Radar point cloud modalities, we use a batch size of $(8, 8, 10)$ and uniformly sample 512 points per frame. All experiments are conducted on a workstation equipped with eight NVIDIA GeForce RTX 3090 GPUs.

\subsection{Single-Modal Recognition}
For baseline evaluations on silhouette and pose modalities, we adopt the original default configurations of each method. For extended analysis, we apply different modeling strategies. For image-based modalities, we employ GaitBase~\cite{fan2023opengait} trained for 60{,}000 iterations using SGD~\cite{robbins1951stochastic}. For pose input, we use GPGait++~\cite{meng2025fastposegait} trained for 40{,}000 iterations with the Adam optimizer~\cite{kingma2014adam}. For point cloud modalities, we adopt LidarGait++~\cite{shen2025lidargait++}, trained for 40{,}000 iterations using SGD~\cite{robbins1951stochastic}.

\subsection{Cross-Modal Retrieval}
Inspired by CL-Gait~\cite{guo2024camera}, we adopt a two-stream architecture for cross-modal gait recognition. We use the GaitBase~\cite{fan2023opengait} framework, where Stage~1 contains modality-specific parameters, while Stages~2-4 share parameters across modalities. Both branches output features of dimension $16 \times 256$.

\vspace{-5mm}
\paragraph{Loss Function.}
We train the network using a combination of symmetric cross-modality triplet loss and cross-entropy loss, equally weighted. To enhance cross-modal feature alignment, we modify the triplet formulation by selecting the anchor from one modality and the positive/negative samples from the other modality:
\begin{equation}
\begin{aligned}
L_{\text{cross-triplet}} = \frac{1}{2} (&
L_{\text{triplet}}(A_{\text{modal1}}, P_{\text{modal2}}, N_{\text{modal2}}) \\
&+\, L_{\text{triplet}}(A_{\text{modal2}}, P_{\text{modal1}}, N_{\text{modal1}})
).
\end{aligned}
\end{equation}

We further apply independent cross-entropy losses to identity predictions from each modality:
\begin{equation}
L_{\text{ce}}^{\text{modal}} = -\sum_{i=1}^{c} y_i \log(\hat{y}_i),
\end{equation}
\begin{equation}
L_{\text{ce}} = \frac{1}{2} \left(
L_{\text{ce}}^{\text{modal1}}
+
L_{\text{ce}}^{\text{modal2}}
\right),
\end{equation}
where $c$ denotes the number of identity classes, $y_i$ is the ground-truth one-hot label, and $\hat{y}_i$ is the predicted probability for class $i$.

The final objective is computed as:
\begin{equation}
L_{\text{total}} = \frac{1}{2} \left( L_{\text{cross-triplet}} + L_{\text{ce}} \right).
\end{equation}

\subsection{Multi-Modal Recognition}

We follow the two-stream fusion strategy proposed in MultiGait++~\cite{jin2025exploring} and investigate several representative visual modality combinations.
Besides, for the fusion of LiDAR point clouds and Radar point clouds, we adopt a dual-stream LidarGait++ architecture without parameter sharing. Each modality is modeled independently, and the resulting features are concatenated before being fed into a shared fully connected layer and BNNeck. The final fused feature has a dimension of $31 \times 256$.
All of the models are trained for 60{,}000 iterations using SGD.

\subsection{Paramers Comparison}
Table~\ref{tab:params_flops} presents the parameter counts and FLOPs for all modality configurations evaluated in our study. The Omni Multi-Modal entry corresponds to our proposed OmniGait model.
Despite supporting every task across the Single-Modal, Cross-Modal, and Multi-Modal settings with a single unified architecture, OmniGait remains remarkably lightweight, requiring only 9.96M parameters. This highlights the efficiency and scalability of our design, enabling broad modality coverage without incurring significant computational overhead.

    
\subsection{Cross-Dataset Evaluation on SUSTech1K}
\begin{table*}[t]
   	\centering
   	\small
   	\setlength{\tabcolsep}{4pt}
   	\begin{tabular}{lcccccccccc}
   		\toprule
   		\multirow{2}{*}{Input Modality} & \multicolumn{8}{c}{Probe Sequence} & \multicolumn{2}{c}{Overall} \\
   		\cmidrule(lr){2-9} \cmidrule(lr){10-11}
   		& Normal & Bag & Clothing & Carrying & Umbrella & Uniform & Occlusion & Night & Rank1 & Rank5 \\
   		\midrule
   		Lidar Depth      & 13.67 & 10.81 & 4.88  & 7.29  & 1.14  & 7.06  & 10.88 & 9.55  & 7.77  & 17.38 \\
   		RGB              & 43.15 & 30.07 & 20.52 & 32.45 & 27.93 & 24.33 & 32.31 & 44.66 & 32.03 & 53.21 \\
   		Silhouette       & 47.24 & 44.68 & 26.28 & 41.78 & 39.42 & 42.04 & 42.98 & 18.08 & 42.65 & 62.15 \\
   		RGB+Silhouette   & 63.93 & 54.39 & 35.98 & 52.71 & 50.98 & 48.86 & 46.97 & 32.86 & 53.07 & 70.96 \\
   		\bottomrule
   	\end{tabular}
\caption{Cross-dataset evaluation on SUSTech1K.
	OmniGait is trained on MMGait and directly evaluated without fine-tuning.}
\label{tab:cross_dataset}
\end{table*}
To further evaluate the generalization capability of OmniGait, we conduct a cross-dataset evaluation by directly transferring the model trained on MMGait to SUSTech1K without any fine-tuning.
This setting is particularly challenging due to significant domain gaps, including differences in sensor configuration, data distribution, environmental conditions, and identity diversity.
All results are obtained under a strict zero-shot transfer protocol.

The detailed cross-dataset results on SUSTech1K are reported in Table~\ref{tab:cross_dataset}.
Under the strict zero-shot transfer setting (trained on MMGait and evaluated without fine-tuning), OmniGait achieves 7.77\% Rank-1 accuracy with LiDAR depth input.
Using RGB input improves the performance to 32.03\%, while silhouette input further boosts it to 42.65\%, indicating that shape-based representations exhibit stronger cross-domain robustness than appearance-only cues.

When fusing RGB and silhouette modalities, the performance is substantially improved to 53.07\% Rank-1 and 70.96\% Rank-5.
The gain from multi-modal fusion is consistent across all probe conditions.
These results demonstrate that OmniGait is capable of learning transferable representations, and multi-modal integration effectively enhances robustness under distribution shift.

\section{Discussion}
The experimental results on MMGait reveal several important observations that offer insights and future directions for multimodal gait recognition research:
\vspace{-5mm}
\paragraph{(1) Cross-modal retrieval remains highly challenging.}
Although certain similar modalities achieve relatively strong cross-modal retrieval performance, most modality pairs still exhibit substantial difficulty, particularly under cross-clothing conditions. A key challenge ahead is how to enable models to simultaneously improve cross-modal alignment and cross-covariate robustness, which remain two conflicting objectives.
\vspace{-5mm}
\paragraph{(2) Multi-modal fusion provides substantial benefits.}
Our results show that the complementary information across modalities is crucial for identity recognition. For example, combining RGB silhouettes with LiDAR projected depth leads to a 19.7\% improvement in challenging cross-clothing scenarios. This improvement stems from the complementary strengths of the two modalities: LiDAR provides stable geometric depth cues, while RGB supplies richer and more discriminative shape information. Their combination significantly enhances robustness under varying appearance conditions.
\vspace{-5mm}
\paragraph{(3) Omni Multi-Modal Recognition presents both opportunities and challenges.}
The Omni Multi-Modal Gait Recognition task is inherently challenging, as it requires a unified framework to handle heterogeneous sensing modalities (e.g., RGB, IR, Depth, LiDAR) while simultaneously supporting diverse retrieval paradigms, including single-modal recognition, multi-modal fusion, and cross-modal retrieval. The large domain gaps across modalities, discrepancies in data distributions, and modality-specific noise patterns make it difficult to learn a shared representation that is both discriminative and modality-invariant. In practice, a unified model often sacrifices single-modal optimality compared to modality-specific counterparts, and its cross-covariate robustness remains constrained under real-world variations.
To establish a feasible starting point for this challenging setting, we introduce OmniGait as a baseline framework. In the current implementation, 3D point cloud data are projected into depth maps before being fed into the network, which helps reduce domain discrepancies between geometric and image-based modalities and enables shared backbone processing. While this design simplifies cross-modal alignment, it inevitably discards part of the intrinsic geometric structure. Directly modeling raw 3D point clouds within a unified omni-modal architecture could therefore be a promising direction for future research, potentially allowing richer geometric cues to be preserved.

\vspace{-5mm}
\paragraph{Limitations:} MMGait does not enforce strict temporal synchronization across modalities, as heterogeneous sensors in real-world deployments naturally differ in sampling rates, exposure cycles, and hardware triggering pipelines. Nevertheless, to approximate synchronization across modalities, we made the following efforts: (1) Frame-Level Synchronization: RGB and Depth are inherently aligned, as they are captured from the same device, ensuring frame-level synchronization for these two modalities. (2) Sequence-Level Synchronization: For other modalities, we perform sequence-level synchronization by recording the start and end timestamps of each device's recording session, enabling approximate temporal alignment across modalities during preprocessing.



\end{document}